\documentclass[lettersize,journal]{IEEEtran}
\usepackage{amsmath,amsfonts}
\usepackage{algorithmic}
\usepackage{algorithm}
\usepackage{array}
\usepackage[caption=false,font=normalsize,labelfont=sf,textfont=sf]{subfig}
\usepackage{textcomp}
\usepackage{stfloats}
\usepackage{url}
\usepackage{verbatim}
\usepackage{graphicx}
\usepackage{cite}
\usepackage{amsfonts}
\usepackage{xcolor}
\usepackage{float}
\usepackage{booktabs}
\usepackage{hyperref}
\usepackage{dsfont}
\usepackage{enumitem}
\usepackage{orcidlink}

\begin{document}

\title{Tractable Probabilistic Models for Investment Planning}

\author{
Nicolás Cuadrado 
\orcidlink{0000-0002-7791-2388},
Mohannad Takrouri 
\orcidlink{0000-0001-8523-3601},
Jiří Němeček 
\orcidlink{0009-0005-0585-1642},
Martin Takáč 
\orcidlink{0000-0001-7455-2025},
Jakub Mareček 
\orcidlink{0000-0003-0839-0691} 
\thanks{N. Cuadrado and M. Takrouri contributed equally to this work.}%
\thanks{N. Cuadrado, M. Takrouri, and M. Takáč are with the Department of Machine Learning, Mohamed bin Zayed University of Artificial Intelligence, Abu Dhabi, UAE (e--mail: \{nicolas.avila, mohannad.takrouri, martin.takac\}@mbzuai.ac.ae).}%
\thanks{J. Němeček and J. Mareček are with the Department of Computer Science, Czech Technical University in Prague, Prague, Czech Republic (e--mail: \{nemecj38, jakub.marecek\}@fel.cvut.cz).}%
}




\maketitle

\newcommand{\TPM}[1]{\text{TPM}(#1)}


\begin{abstract}
Investment planning in power utilities, such as generation and transmission expansion, requires decisions under substantial uncertainty over decade--long horizons for policies, demand, renewable availability, and outages, while maintaining reliability and computational tractability. Conventional approaches approximate uncertainty using finite scenario sets (modeled as a mixture of Diracs in statistical theory terms), which can become computationally intensive as scenario detail increases and provide limited probabilistic resolution for reliability assessment. We propose an alternative based on tractable probabilistic models, using sum--product networks (SPNs) to represent high--dimensional uncertainty in a compact, analytically tractable form that supports exact probabilistic queries (e.g., likelihoods, marginals, and conditionals). This framework enables the direct embedding of chance constraints into mixed--integer linear programming (MILP) models for investment planning to evaluate reliability events and enforce probabilistic feasibility requirements without enumerating large scenario trees. We demonstrate the approach on a representative planning case study and report reliability--cost trade--offs and computational behavior relative to standard scenario--based formulations.
\end{abstract}



\section*{Nomenclature}
\addcontentsline{toc}{section}{Nomenclature}
{\fontsize{9.5}{10}\selectfont
\begin{IEEEdescription}[\IEEEusemathlabelsep\IEEEsetlabelwidth{$X = \{x_1,\dots,x_D\}$}]
\item[$X = \{x_1,\dots,x_D\}$] Investment decision variables.
\item[$\Xi$] Random vector over horizon $T$.
\item[$\hat{\Xi}$] Estimate of $\Xi$.
\item[$\xi$] Realization of uncertainty w.r.t. $\hat{\Xi}$.
\item[$Y$] Adequacy indicator.
\item[$G(x,\xi)$] System adequacy constraint.
\item[$C^{\mathrm{inv}}(x)$] Investment cost.
\item[$C^{\mathrm{op}}(x,\xi)$] Operational cost under $\xi$.
\item[$C^{\mathrm{total}}(x)$] Total cost (inv. + expected op.).
\item[$p_{\Xi}(\xi)$] Density of stochastic variables.
\item[$p_{\Xi,Y,X}(\xi,y,x)$] Joint density of $(\Xi,Y,X)$.
\item[$p_X(x)$] Marginal density of $X$.
\item[$P_{\xi}(G(x,\xi)\ge0)$] Chance constraint probability.
\item[$\varepsilon$] Adequacy violation probability.
\item[$\mathrm{TPM}(\cdot)$] Tractable probabilistic model.
\item[$\mathcal{V}, \mathcal{E}$] Nodes and edges of SPN graph.
\item[$\mathcal{V}_\Sigma, \mathcal{V}_\Pi, \mathcal{V}_L$] Sum, product, and leaf node sets.
\item[$O_n(x_{\psi(n)})$] Output of node $n$.
\item[$\psi(n)$] Scope (variables covered by node $n$).
\item[$w_{a,n_\Sigma}$] Weight of child $a$ in sum node $n_\Sigma$.
\item[$ch(n)$] Set of children of node $n$.
\item[$o_n$] Log--probability output of node $n$.
\item[$m_{a,n_\Sigma}$] Binary variable for active child.
\item[$\mathrm{T}^{\mathrm{LL}}_{n_\Sigma}$] Big--M constant for sum node.
\item[$K$] Number of scenarios.
\end{IEEEdescription}
}
\section{Introduction}

\IEEEpubidadjcol 

Climate change impacts intensify, and policymakers are accelerating the transition toward cleaner power systems. While renewables provide a feasible pathway to decarbonization and meeting greenhouse gas reduction targets, their weather--dependent and stochastic nature introduces profound operational challenges \cite{paragElectricityMarketDesign2016}. Reliability is paramount: supply interruptions can propagate through the economy, disrupt critical infrastructure, and undermine social welfare. At the same time, the variability of renewable sources increases the need for additional reserves, flexibility options, and grid reinforcements, all of which carry high costs that grow with penetration levels \cite{edenhoferRenewableEnergySources2011, impramChallengesRenewableEnergy2020}. 

Power grid operators must balance affordability, reliability, and resilience when making infrastructure investments, making rigorous planning essential. Capacity expansion planning (CEP) determines what capacity to build, when to build it, and which technologies to deploy over multi--decade horizons under significant uncertainty. Originally developed in vertically integrated systems prior to market liberalization, CEP frameworks aimed to expand generation and transmission infrastructure at minimum cost \cite{conejoInvestmentElectricityGeneration2016}. In modern systems, this methodology extends to the new market dynamics and operational conditions brought by the increasing penetration of solar, storage, and demand--side management strategies. It also depends on the power system segment (generation \cite{oderinwaleInvestigatingImpactFlexible2020}, transmission \cite{taylorConvexOptimizationPower2015, loschenbrandTransmissionExpansionModel2021, gbadamosiMultiperiodCompositeGeneration2020}, and distribution \cite{xieExpansionPlanningActive2020, gbadamosiMultiperiodCompositeGeneration2020}) and varies across time horizons, and decision variables \cite{mottaSurveyOptimizationModels2024}, being Generation Capacity Expansion Planning (GEP) and Transmission Network Capacity Expansion Planning (TNEP) the most prominent in the literature.

A dominant paradigm in this domain is stochastic programming (SP), as explained in \cite{barrosoDecisionMakingUncertainty2006, tahananLargescaleUnitCommitment2015a, vanackooijLargescaleUnitCommitment2018a}. These works concisely motivate the challenges of using SP for capacity expansion planning. First, they argue that simply removing stochasticity typically yields solutions that are neither good nor even feasible. Second, they outline common strategies to mitigate the ``curse of dimensionality'' in SP: scenario reduction methods that trade accuracy for tractability, and decomposition--based techniques that prioritize tractability. Recent research in \cite{zamparaCapacityExpansionPlanning2025} supports this view on strategies and emphasizes that hybrid approaches are also valid. Additionally, the seminal book referenced in \cite{conejoInvestmentElectricityGeneration2016} consolidates widely used SP--based methods in GEP and TNEP.

Scenario--based stochastic CEP remains prevalent but can incur significant computational expense as the number of scenarios and the degree of modeled uncertainty increase \cite{zakariaUncertaintyModelsStochastic2020}, motivating tractability--focused methods such as scenario reduction, partitioning, and decomposition. To reduce the scenario space, some works focus on selecting scenarios, such as \cite{sunDataDrivenRepresentativeDay2019}, which presents a data--driven methodology to choose representative periods (e.g., days). In contrast, scenario partitioning can approximate multi--scale uncertainty by clustering scenarios into groups that share common recourse decisions while retaining the complete set of scenarios \cite{zhaoScenarioPartitioningMethods2021}. Decomposition--based approaches target scalability in large--scale stochastic CEP formulations; for example, \cite{pecciRegularizedBendersDecomposition2025} develops a regularized Benders Decomposition framework that leverages distributed computing to accelerate convergence in large--scale CEP and enables high--resolution capacity expansion models while explicitly modeling stochastic operational scenarios. Surveys and monographs highlight that computational tractability is a key challenge across stochastic and robust expansion--planning formulations, including those with explicit reliability considerations \cite{choRecentAdvancesChallenges2022,conejoInvestmentElectricityGeneration2016}.

Robust expansion planning, unlike scenario--based planning, seeks a solution that remains feasible and efficient across a wide range of possible, even unfavorable, future scenarios. In the field of TNEP, Jabr \cite{jabrRobustTransmissionNetwork2013a} proposed a method that uses uncertainty sets to eliminate the need to know the distribution of net injection uncertainty, providing a robust exact solution for all realizations within the set. Using a similar resource, Ruiz and Conejo \cite{ruizRobustTransmissionExpansion2015} proposed a bi--level MILP model for TNEP problems and solved it via decomposition with a cutting--plane algorithm. AC--based TNEP is computationally challenging, underscoring the tractability cost of physics--rich mixed--integer formulations. Ghaddar and Jabr \cite{jabrPowerTransmissionNetwork2019} proposed a method that builds on the semidefinite relaxation of the AC optimal power flow problem (ACOPF) to solve the TNEP problem with AC--power flow constraints using a branch--and--bound algorithm. These works motivate the need for tractable uncertainty representations in long--horizon planning.

Recent multi--period expansion models link long--horizon investment with short--term operational uncertainty. For instance, \cite{rintamaki_achieving_2024} proposes a simultaneous solution to GEP and TNEP using a tri--level model that combines multi--year investment horizons with robust demand realization and stochastic multi--hour operational scenarios. While such formulations improve temporal realism, uncertainty is still handled through scenario and worst--case formulations rather than an explicit learned joint distribution. 

Classical chance--constrained planning evaluates expectations and violation probabilities under a predefined scenario structure, implicitly assuming a fixed probability distribution of uncertainty \cite{wangStochasticUnitCommitment2013,chenChanceconstrainedMultistageStochastic2024}. In practice, distributional information is inferred from limited historical data and may be misspecified, motivating distributionally robust formulations that optimize over a family of plausible distributions. Distributionally robust generation expansion planning (DRO--GEP) builds upon that motivation by modeling renewable uncertainty through a moment--based ambiguity set \cite{pourahmadi_distributionally_2020,pourahmadi_distributionally_2021}. Although DRO frameworks also encompass metric--based ambiguity sets (e.g., Wasserstein), such formulations yield large MILP reformulations due to the additional decision stages introduced to capture uncertainty, and the resulting increase in variables and constraints, which grows with the number of samples \cite{chenWassersteinDistancebasedDistributionally2022}.

Moment--based sets remain particularly attractive in capacity expansion planning due to their analytical tractability and compatibility with mixed--integer conic reformulations. The chance--constrained formulation in \cite{pourahmadi_distributionally_2020} minimizes expected investment using a hybrid model that combines scenario--based planning for long--term uncertainty with ambiguity sets to model short--term uncertainty. The latter are defined by a family of distributions with the same first-- and second--order moments derived from historical data. This method provides protection against distributional misspecification while remaining less conservative than classical robust optimization. Pourahmadi and Kazempour \cite{pourahmadi_distributionally_2021} proposed a DRO--GEP framework that constructs moment--based ambiguity sets that take advantage of the unimodality of wind power production, and models operational--limit violation risk in two alternative forms: a chance--constrained formulation and a CVaR--constrained formulation. Nevertheless, these moment--based DRO approaches evaluate uncertainty only through worst--case summary statistics and do not learn an explicit joint probability density. At the operational level, related DRO formulations have been developed for power flow optimization using CVaR--based risk measures \cite{jabrDistributionallyRobustCVaR2020}, reinforcing the tractability of moment--based ambiguity modeling in power system applications.

Progress in probabilistic modeling has advanced rapidly in recent years. Methods like generative adversarial networks \cite{goodfellowGenerativeAdversarialNetworks2014}, variational autoencoders \cite{kingmaAutoEncodingVariationalBayes2022}, normalizing flows \cite{papamakariosNormalizingFlowsProbabilistic2021}, and diffusion models \cite{ho_denoising_2020} approximate complex data distributions with high fidelity. However, as Choi \cite{choiProbabilisticCircuitsUnifying2020} highlighted, these models improve at fitting data distributions but become less reliable for probabilistic reasoning, undermining their role in principled decision--making. That is why the aforementioned methods in scenario--based and distributionally robust formulations resort to probabilistic models (for uncertainty) compatible with widely used optimization techniques, sacrificing expressivity for exactness.

Tractable probabilistic models (TPMs) are an alternative family of models that represent complex, high--dimensional probability distributions in a compact form. Choi \cite{choiProbabilisticCircuitsUnifying2020} unifies a broad family of TMPs under the name of probabilistic circuits (PC) and defines them as rooted DAGs with sum--product structure (mixtures and factorizations), enabling exact probabilistic queries such as marginals and conditionals in time linear in circuit size under standard structural conditions.

TMPs rely on the foundational compilation results that show how arithmetic circuits enable efficient computation of large classes of probabilistic queries via circuit evaluations and derivatives \cite{darwicheLogicalApproachFactoring2002a}. Building on this representation paradigm, models such as sum--product networks (SPNs) \cite{poonSumproductNetworksNew2011a}, probabilistic sentential decision diagrams (PSDDs) \cite{kisaProbabilisticSententialDecision2014}, and probabilistic generating circuits (PGCs) \cite{zhangProbabilisticGeneratingCircuits2021} learn explicit joint distributions while maintaining tractable inference guarantees. This property is attractive for expansion planning under deep uncertainty, where reliable decisions require scalable probabilistic reasoning \cite{lavenezianaCriticalReviewEnergy2023a}. 

Recent work shows that probabilistic circuits, specifically SPNs, can be embedded within mixed--integer optimization by formulating SPN likelihood computations directly as MILP constraints, enabling likelihood--based feasibility requirements \cite{nemecek_generating_2025}. SPNs \cite{poonSumproductNetworksNew2011a} represent joint distributions as computational graphs in which sum nodes encode mixtures and product nodes encode factorizations. Under structural constraints (e.g., completeness and consistency), they provide a particularly natural instantiation for optimization--embedded probabilistic reasoning. In the CEP case, this enables direct evaluation of reliability events within an optimization framework, without resorting to sampling--based inference.

The literature above shows that CEP is commonly addressed through (i) scenario--based stochastic programming, (ii) robust optimization over uncertainty sets, and (iii) distributionally robust optimization over ambiguity sets, often combined with chance/CVaR risk constraints and decomposition techniques to improve tractability. In contrast, TPMs represent uncertainty densities in a structured form that supports polynomial--time inference, enabling probabilistic constraints to be evaluated or enforced within mixed--integer optimization. This provides an alternative pathway to integrate uncertainty without explicit scenario trees or complex ambiguity--set reformulations. In this work, we propose an alternative paradigm for solving CEP, in which uncertainty is modeled by a tractable probabilistic model and reliability is enforced via probabilistic inference embedded in the optimization process. 

We adopt a sum--product network (SPN) to learn an explicit joint distribution over key uncertainties and to evaluate reliability events via exact, tractable queries within a mixed--integer CEP formulation. This shifts computation from scenario enumeration and ambiguity--set reformulations to offline SPN learning and repeated tractable inference in the optimization loop. This framework is adaptable across capacity expansion settings; the computed reliability is exact \emph{given the learned model} and is compatible with the worst--case distributional guarantees of DRO approaches. The main contributions of this paper are:

\begin{itemize}
    \item \textbf{TPM--based CEP formulation:} We propose a novel reformulation of the capacity expansion planning problem in which reliability constraints are expressed and evaluated via exact probabilistic queries of a tractable probabilistic model, rather than via discrete scenarios or worst--case ambiguity sets.
    \item \textbf{SPN--based joint uncertainty model:} We develop an SPN--based uncertainty model capturing mixed continuous--discrete drivers, enabling explicit dependency modeling while preserving exact tractable inference.
    \item \textbf{MILP integration of probabilistic inference:} We show how SPN--based probability evaluation can be integrated into a mixed--integer CEP model, enabling direct enforcement of probabilistic feasibility/reliability requirements within the optimization loop.
    \item \textbf{Empirical demonstration on a representative case study:} We provide a GEP case study illustrating feasibility, computational behavior, and reliability--cost trade--offs under the proposed inference--based planning paradigm.
\end{itemize}
\section{Preliminaries} \label{sec:preliminaries}
We first introduce a reference expansion planning formulation that serves as a common baseline for the methods developed in the remainder of the paper. The goal is not to prescribe a particular application, but to define a generic investment--operation structure, specify the uncertainty objects, and establish the notation used throughout. Although the notation allows generation, transmission, and storage decisions, for generality, the experiments in this paper focus on \emph{generation} expansion planning (GEP).
\subsection{Investment Decisions}
Capacity expansion formulations generally consist of (here--and--now) investment decisions and second--stage (wait--and--see) operational decisions \cite{doi:10.1137/1.9781611973433.ch2,bounitsisDatadrivenScenarioGeneration2022}. For the reference case, let $X=\{x_1,\dots,x_D\}$ denote the set of investment decisions, where $D$ is the number of available decisions. For instance, given a set of generation units $\mathcal{G} = \{g_1,\dots,g_{G}\}$, a set of transmission lines $\mathcal{L} = \{\ell_1,\dots,\ell_{L}\}$, and a set of storage units $\mathcal{S}=\{s_1,\dots,s_{S}\}$, a single decision could have the form
\begin{align}
    x_d = \Big\{
        & G,g_1^{\mathrm{cap}},\dots,g_{G}^{\mathrm{cap}}, 
        L,\ell_1^{\mathrm{cap}},\dots,\ell_{L}^{\mathrm{cap}}, \notag\\
        & S,s_1^{\mathrm{cap}},\dots,s_{S}^{\mathrm{cap}}, 
        \dots
    \Big\},
    \label{eq:decision_vars}
\end{align}
specifying the number and capacities of generation units ($G$), transmission lines ($L$), and storage units ($S$). Decisions may also include portfolio or technology adoption choices, such as binary or continuous variables indicating technology selection or minimum/maximum technology shares. 

\subsection{Stochastic Realizations}

The second group represents short--term operational decisions driven by stochastic factors, such as weather and demand, over a horizon of length $T$. We define $\Xi=\{\xi_1\oplus\dots\oplus\xi_T\}$ as the concatenation of all stochastic variables after uncertainty realization. A realization at time step $t$ is defined as:
\begin{align}
    \xi_t = \Big\{ 
        & 
        g^{\mathrm{out}}_{1,t},\dots,g^{\mathrm{out}}_{G,t},
        g^{\mathrm{curt}}_{1,t},\dots,g^{\mathrm{curt}}_{G,t},
        g^{\mathrm{res}}_{1,t},\dots,g^{\mathrm{res}}_{G,t}, 
        \notag\\
        &
        \ell^{\mathrm{P}}_{1,t},\dots,\ell^{\mathrm{P}}_{L,t},
        \ell^{\mathrm{Q}}_{1,t},\dots,\ell^{\mathrm{Q}}_{L,t},
        s^{\mathrm{ch}}_{1,t},\dots,s^{\mathrm{ch}}_{S,t},
        \notag\\
        &
        s^{\mathrm{dch}}_{1,t},\dots,s^{\mathrm{dch}}_{S,t},
        \dots
    \Big\},
    \label{eq:stochastic_vars}
\end{align}
including generation output ($g^{\mathrm{out}}_{1,t},\dots,g^{\mathrm{out}}_{G,t}$), curtailment ($g^{\mathrm{curt}}_{1,t},\dots,g^{\mathrm{curt}}_{G,t}$), reserve decisions ($g^{\mathrm{res}}_{1,t},\dots,g^{\mathrm{res}}_{G,t}$), operational values on the power lines such as active power ($\ell^{\mathrm{P}}_{1,t},\dots,\ell^{\mathrm{P}}_{L,t}$) and reactive power ($\ell^{\mathrm{Q}}_{1,t},\dots,\ell^{\mathrm{Q}}_{L,t}$), charging ($s^{\mathrm{ch}}_{1,t},\dots,s^{\mathrm{ch}}_{S,t}$) and discharging ($s^{\mathrm{dch}}_{1,t},\dots,s^{\mathrm{dch}}_{S,t}$) decisions, among others.

\subsection{One--Stage Generic Scenario Expansion Problem}
Together, $X$ and $\Xi$ embed uncertainty into the investment planning, whose objective is to minimize cost while maintaining reliability. We will consider the second--stage decisions and the stochastic realizations of the operation to limit this problem to investment decisions. As the space of possible realizations $\xi_t$ is large, exhaustive evaluation is intractable \cite{zakariaUncertaintyModelsStochastic2020}. Instead, an estimate $\hat{\Xi}$ of representative cases is constructed, and $x$ is optimized over each $\xi \sim \hat{\Xi}$ as follows:
\begin{align}
    \min_{x}\quad 
    &
    C^{\mathrm{inv}}(x) +
    \mathbb{E}_{\xi \sim \hat \Xi} \left[
        C^{\mathrm{op}}(x,\xi)
    \right]
    \label{eq:scenario_expansion_og}
    \\
    \text{s.t.}\quad 
    & G(x,\xi) \geq 0,
    \notag\\
    & g(x) = 0, 
    \notag\\
    & h(x) \leq 0 \notag
    ,
\end{align}
to posteriorly pick the least expensive solution among the selected scenarios. The constraints in \eqref{eq:scenario_expansion_og} capture operational and policy requirements that restrict the investment objective. This compact form encompasses diverse system constraints without explicit enumeration.
\section{Methodology}
\subsection{Chance--Constrained Capacity Expansion Problems} \label{ssec:cc_expansion}

The traditional CEP described in \autoref{sec:preliminaries} relies on finding a global optimum of an approximated problem based on an estimate $\hat \Xi$ built from a set of representative realizations of the uncertainty, effectively modeling the uncertainty as a mixture of Dirac distributions. Formulation \eqref{eq:scenario_expansion_og} seeks the least--cost decision $x$ for which all sampled scenarios $\xi \sim \hat{\Xi}$ satisfy the system adequacy constraints simultaneously (possibly up to a predefined adequacy level). When uncertainty is high, this requirement leads to a significant computational burden. An alternative is to assume access to a probabilistic model of $\mathbb{P}_{\xi \sim p(\Xi)} \left(G(x,\xi) \geq 0\right)$, which makes it possible to formulate a chance--constrained optimization problem, where constraint satisfaction is required only with a prescribed violation probability threshold $\varepsilon$:
\begin{align}
    \min_{x}\quad 
    & 
    C^{\mathrm{inv}}(x) + 
    \mathbb{E}_{\xi \sim p(\Xi)} \left[C^{\mathrm{op}}(x,\xi)\right]
    \label{eq:scenario_expansion_cc}\\
    \text{s.t.}\quad 
    & \mathbb{P}_{\xi \sim p(\Xi)} \left(G(x,\xi) \geq 0\right) \ge 1-\varepsilon,
    \notag\\
    & g(x) = 0, \notag\\
    & h(x) \le 0. \notag
\end{align}

\subsection{TPM--Based Chance--Constrained Capacity Expansion Problems}

Solving \eqref{eq:scenario_expansion_cc} requires (i) a reliable probabilistic model of the chance constraint and (ii) being able to perform inference over it many times. TPMs address both needs by enabling polynomial--time inference and compatibility with optimization pipelines, as shown in \cite{nemecek_generating_2025}. To enable the integration in such a framework, we start by defining a binary random variable $Y=\mathds{1}(G(x,\xi) \geq 0)$ to express the chance constraint in probabilistic form:
\begin{align}
    \mathbb{P}_{\xi \sim p(\Xi)} \left(G(x,\xi) \geq 0\right)
    & = \mathbb{P}_{\xi \sim p(\Xi),x \sim X} \left(G(x,\xi) \geq 0|X=x\right) \notag \\
    & = \int_{-\infty}^{\infty} p_{\Xi\mid X}(\xi | x) \mathds{1}(G(x,\xi) \geq 0) d\xi \notag \\ 
    & = \int_{-\infty}^{\infty} p_{\Xi,Y\mid X}(\xi,1 | x) d\xi \notag \\
    & = \int_{-\infty}^{\infty} \frac{p_{\Xi, Y, X}(\xi, 1, x)}{p_{X}(x)} d\xi
    .
    \label{eq:cc_redef}
\end{align}

The reformulation \eqref{eq:cc_redef} shows that evaluating the chance constraint reduces to estimating joint and marginal probability densities over the random variables $(X,\Xi,Y)$. Because TPMs support polynomial--time inference, these quantities can be computed directly through appropriate density queries:
\begin{align}
    \TPM{\xi,y,x}
    & \approx p_{\Xi, Y, X}(\xi, y, x) 
    \label{eq:tpm_joint_xyxi}
    ,\\
    \TPM{y,x}
    & \approx p_{Y, X}(y, x)
    \label{eq:tpm_joint_xy}
    ,\\
    \TPM{x} 
    & \approx p_{X}(x)
    \label{eq:tpm_joint_x},
\end{align}
where $y \sim Y$ the adequacy outcome indicating whether the system meets demand ($y=1$) or experiences a shortfall ($y=0$).

Depending on the query, the TPM can estimate the joint distribution of variables as in \eqref {eq:tpm_joint_xyxi} and \eqref{eq:tpm_joint_xy}, or compute the marginal of one of the variables as in \eqref{eq:tpm_joint_x} for $X$. Using a TPM, we can reformulate the chance constraint as
\begin{equation}
    \frac{\TPM{x,y=1}}{\TPM{x}} \ge 1-\varepsilon.
    \label{eq:tpm_approx_cc}
\end{equation}

In other terms, \eqref{eq:tpm_approx_cc} uses an arbitrary number of samples $(x,\xi)$ to learn a joint distribution over $(X,\Xi,Y)$. Once learned, marginalization and conditioning enable direct evaluation of the required probabilities. Knowing the distribution of $x$, $p_{X}(x)$ can replace the denominator in \eqref{eq:tpm_approx_cc} directly.

\subsection{Optimizing TPM--Based Chance Constraint}

Large--scale energy planning and operational problems are typically formulated as linear or mixed--integer linear programs (LP/MILP) due to their computational scalability and compatibility with commercial solvers. Applications such as GEP, TNEP, unit commitment, and market clearing rely on these formulations \cite{conejoInvestmentElectricityGeneration2016}. 

The presented chance--constrained formulation is MILP--compatible, following the method described in \cite{nemecek_generating_2025}, enabling linear encoding of SPNs. 

An SPN represents a joint distribution as a composition of weighted sums and products of tractable components. Formally, it is a probabilistic circuit defined by a tuple $(\mathcal{G}, \psi, \theta)$, where $\mathcal{G} = (\mathcal{V}, \mathcal{E})$ is a directed acyclic graph with vertex set $\mathcal{V}$ and edge set $\mathcal{E}$. The function $\psi: \mathcal{V} \rightarrow 2^{[P]}$ assigns a scope to each node, and $\theta$ denotes the model parameters. For each node $n \in \mathcal{V}$, let $\mathrm{ch}(n)$ denote its children and $x_{\psi(n)}$ the variables in its scope.

A \emph{leaf node} $n^{L} \in \mathcal{V}^{L} = \{n \mid \mathrm{ch}(n) = \emptyset\}$ represents a probability distribution over its scope,
\begin{align}
    O_{n^{L}}(x_{\psi(n^{L})}) = p(x_{\psi(n^{L})}; \theta_{n^{L}}),
    \label{eq:leaf_node}
\end{align}
where $p(\cdot)$ is a tractable distribution parameterized by $\theta_{n^{L}}$. 

A \emph{product node} $n^{\Pi} \in \mathcal{V}^{\Pi}$ computes the product of its children's output,
\begin{align}
    O_{n^{\Pi}}(x_{\psi(n^{\Pi})}) = 
    \prod_{a \in \mathrm{ch}(n^{\Pi})} O_{a}(x_{\psi(a)}).
    \label{eq:prod_node}
\end{align}

To ensure tractability, the scopes of the children of a product node must be \emph{decomposable}, meaning that they are disjoint and collectively cover the parent’s scope:
\begin{align}
    \bigcup_{a \in \mathrm{ch}(n^{\Pi})} \psi(a) & = \psi(n^{\Pi}), \\
    \bigcap_{a \in \mathrm{ch}(n^{\Pi})} \psi(a) & = \emptyset.
    \label{eq:decomposability}
\end{align}

A \emph{sum node} $n^{\Sigma} \in \mathcal{V}^{\Sigma}$ computes a weighted mixture of the distributions defined by its children's output:
\begin{align}
    O_{n^{\Sigma}}(x_{\psi(n^{\Sigma})}) 
    = \sum_{a \in \mathrm{ch}(n^{\Sigma})} 
      w_{a,n^{\Sigma}} \, O_{a}(x_{\psi(a)}),
    \label{eq:sum_node}
\end{align}
where $w_{a,n^{\Sigma}} \ge 0$ and $\sum_{a \in \mathrm{ch}(n^{\Sigma})} w_{a,n^{\Sigma}} = 1$. All children of a sum node must share identical scopes (a property known as \emph{completeness} or \emph{smoothness}):
\begin{align}
    \psi(a_1) = \psi(a_2), \qquad \forall \, a_1, a_2 \in \mathrm{ch}(n^{\Sigma}).
\end{align}

These structural constraints guarantee that SPNs support exact marginalization and conditioning in linear time with respect to the network size. Consequently, SPNs can efficiently compute the densities and expectations required in stochastic and chance--constrained optimization. To integrate an SPN within an LP/MILP problem, its computation must be expressed through linear constraints, which is non--trivial.

Following \cite{nemecek_generating_2025}, leaf nodes $n^{L} \in \mathcal{V}^{L}$ represent tractable univariate distributions. For discrete variables, indicator variables $d_{j,k}$ encode the active category $k$ of feature $j$. For continuous variables, piecewise--linear approximation (e.g., histograms or linearized densities) represents the log--likelihood within the linear model. The corresponding log--space outputs are denoted by $o_{n^{L}}$.

Product node ($n^{\Pi} \in \mathcal{V}^{\Pi}$) computations are performed in logarithmic space, which simplifies their formulation and improves numerical stability, aggregating the log--probabilities of its children additively. Let $\mathrm{ch}(n^{\Pi})$ denote the set of predecessors of node $n^{\Pi}$, each producing an output $o_{a}$. The product node satisfies
\begin{align}
    o_{n^{\Pi}} = \sum_{a \in \mathrm{ch}(n^{\Pi})} o_{a}, 
    \qquad \forall n^{\Pi} \in \mathcal{V}^{\Pi}.
\end{align}

This relation is linear and therefore directly compatible with LP formulations. A sum node $n^{\Sigma} \in \mathcal{V}^{\Sigma}$ computes a weighted mixture of its children’s outputs. In probability space, this corresponds to a convex combination 
\begin{align}
\sum_{a \in \mathrm{ch}(n^{\Sigma})} w_{a,n^{\Sigma}} \, \exp(o_{a}),
\end{align}
which becomes nonlinear in log--space. To preserve tractability, we approximate the logarithm of the sum by its dominant term using a ``log--sum--exp'' upper bound, which can be expressed as
\begin{align}
    o_{n^{\Sigma}} \approx 
    \max_{a \in \mathrm{ch}(n^{\Sigma})} 
    \left( o_{a} + \log w_{a,n^{\Sigma}} \right).
\end{align}

This approximation can be linearized using binary variables $m_{a,n^{\Sigma}} \in \{0,1\}$ indicating which child is active. For each sum node $n^{\Sigma} \in \mathcal{V}^{\Sigma}$ and each child $a \in \mathrm{ch}(n^{\Sigma})$, we define:
\begin{align}
    o_{n^{\Sigma}} &\le o_{a} + \log w_{a,n^{\Sigma}} + m_{a,n^{\Sigma}} \, \text{T}^{\mathrm{LL}}_{n^{\Sigma}}.
    \label{eq:spn_sum_linearization}
\end{align}

Here $T^{\mathrm{LL}}_{n^{\Sigma}}$ is a large constant (big-$M$) ensuring that only one child can determine the active maximum, achieved through the following additional constraint:
\begin{align}
    \sum_{a \in \mathrm{ch}(n^{\Sigma})} m_{a,n^{\Sigma}} &= 
    |\mathrm{ch}(n^{\Sigma})| - 1,
    &\forall n^{\Sigma} \in \mathcal{V}^{\Sigma},
    \label{eq:spn_sum_selector}
\end{align}
yielding a lower bound on the true log--sum--exp value, which is sufficient when maximizing likelihood or probability mass. 

Overall, these constraints encode the SPN structure within a mixed--integer linear program. Product nodes remain linear, while sum nodes require a small number of binary variables per mixture component. This formulation enables the evaluation of likelihood--based or chance--constrained objectives to be handled by standard MILP solvers.

\subsection{Computational Scaling and Scenario Independence}

In traditional scenario--based approaches, the number of variables and constraints grows proportionally to the number of sampled realizations, $\mathcal{O}(K)$, as each scenario introduces additional operational constraints. This significantly increases solver time, especially when thousands of scenarios are required to capture high--dimensional uncertainty. In contrast, the proposed TPM--based formulation decouples optimization from explicit scenario enumeration. Once trained, the SPN evaluates the probabilities required for the chance constraint through a fixed number of inference queries, each computable in linear time in the network size and independent of $K$. Thus, probabilistic inference replaces scenario enumeration, enabling the optimizer to reason over the full uncertainty distribution. Moreover, because TPMs can be trained on arbitrarily large datasets, increasing $K$ improves model accuracy rather than computational burden.

\subsection{TPM--Based Pipeline for Probabilistic Investment Planning}

To demonstrate that TPMs can solve CEP problems, we propose a pipeline organized into three layers around a GEP problem. The first is a realistic simulator that generates large numbers of daily scenarios by combining design decisions (e.g., hardware specifications) with prototypical load profiles and stochastic environmental variables, labeling each realization as adequate or shortfall. The second performs probabilistic learning using SPNs to model the conditional adequacy probability given a design. The third integrates these probabilistic outputs into an MILP solver, enabling optimization through chance constraints on adequacy probability.

\subsubsection{Generation Design Simulator}
The first layer allows for the simulation of investment decisions within a GEP, formalized as a mapping:
\begin{align}
    f : (x, \xi) \;\mapsto\; y,
    \label{eq:simulator_model}
\end{align}
where $x \sim X$ denotes the vector of decision variables (e.g., solar capacity, storage capacity, inverter technology, and storage operation policy), $\xi \sim \Xi$ the realization of relevant stochastic variables (e.g., irradiance, temperature, and load profile), and $y \sim Y$ the adequacy outcome indicating whether the system meets demand ($y=1$) or experiences a shortfall ($y=0$). 

Given a configuration $x$ and a realization of the stochasticity $\xi$, the simulator computes the system’s power balance across a planning horizon, accounting for generation, storage operation, and load demand.
By repeatedly evaluating $f(x,\xi)$ over multiple draws of $\xi$, the simulator produces a dataset of labeled daily realizations:
\begin{align}
    \mathcal{D} = \{(x_i, \xi_i, y_i)\}_{i=1}^{N}.
\end{align}

The resulting dataset provides a statistical foundation for probabilistic learning in Layer 2.

\subsubsection{SPN Training and Inference}
The second layer learns a TPM (specifically an SPN) from the labeled dataset generated in Layer 1 to approximate the joint density $p(X, \Xi, Y)$. Once trained, the SPN supports exact inference, including conditional probabilities such as $\mathbb{P}(Y=1\mid X=x)$, and relevant marginals. Because inference is polynomial in the network size, chance constraints can be evaluated over arbitrarily large uncertainty spaces without explicit scenario enumeration.

\subsubsection{MILP Integration and Optimization}
The third layer integrates the trained SPN into an MILP setup. The chance constraint enforces ``adequacy'', i.e., the event that demand is met up to a specified tolerance. Let $\Delta(x,\xi)$ denote the instantaneous shortfall under decision $x$ and realization $\xi$, and let $\tau \ge 0$ be a prescribed shortfall tolerance. The optimization seeks a configuration $x$ that minimizes total investment and operational costs while ensuring that the adequacy probability remains at $1-\varepsilon$. For simplicity, we restrict the objective to investment cost only, although SPN also enables the evaluation of expectations:
\begin{align}
    \min_{x}\quad & C^{\mathrm{inv}}(x)
    \label{eq:practical_objective}\\
    \text{s.t.}\quad & \mathbb{P}_{\xi \sim p(\Xi)}\left(\Delta(x,\xi) \le \tau\right) \ge 1-\varepsilon.
    \notag
\end{align}

The SPN evaluates the probability term directly via its encoding as linear and mixed--integer constraints, as described in \autoref{ssec:cc_expansion}. This formulation enables standard MILP solvers to compute feasible and near--optimal solutions without Monte Carlo sampling. By combining tractable probabilistic inference with deterministic optimization, the framework selects cost--efficient designs that satisfy adequacy constraints.
\section{Experimental Results}

\subsection{Probability Landscape Comparison} \label{sec:spn_fitting}

To assess the ability of SPN--based methods to learn the uncertainty distribution over system configurations, we compare their estimated probability landscapes with the empirical estimation of the adequacy probability \eqref{eq:practical_objective} across different levels of data availability. The available SPN--based methods are the following:

\subsubsection{SPN} Uses the trained SPN to directly evaluate adequacy probabilities based on the sampled training data for each iteration. Although not MILP--compatible, it provides a reference benchmark for the best achievable SPN--based solution.

\subsubsection{SPN--Max} Implements the MILP--compatible approximation proposed in \cite{nemecek_generating_2025}, replacing log--sum--exp operations with a max--based formulation. This enables embedding SPNs within a tractable optimization framework while enforcing chance constraints.

\subsubsection{SPN--Piecewise} An improved MILP formulation that approximates exponential and logarithmic functions using piecewise--linear segments. For each sum node $n^{\Sigma}$, the maximal input value is subtracted for numerical stability. It approximates exponential components on the range $[\ln 0.001, 0]$, where values are numerically relevant (values below 0.001 are rounded to 0). Then, a logarithm approximation of the sum of exponential components is taken on [1, $|\mathrm{ch}(n^{\Sigma})|$] to cover all possible values. Each function uses five linear segments. The maximal value is added back to obtain a tighter approximation of the true log--sum--exp.

We examine whether high--probability regions, ridges, and exclusion zones are correctly identified, how estimates evolve as additional data are provided (convergence behavior), and where systematic biases or overconfidence may persist. 
\begin{figure*}[tb]
    \centering
    \includegraphics[width=\linewidth]{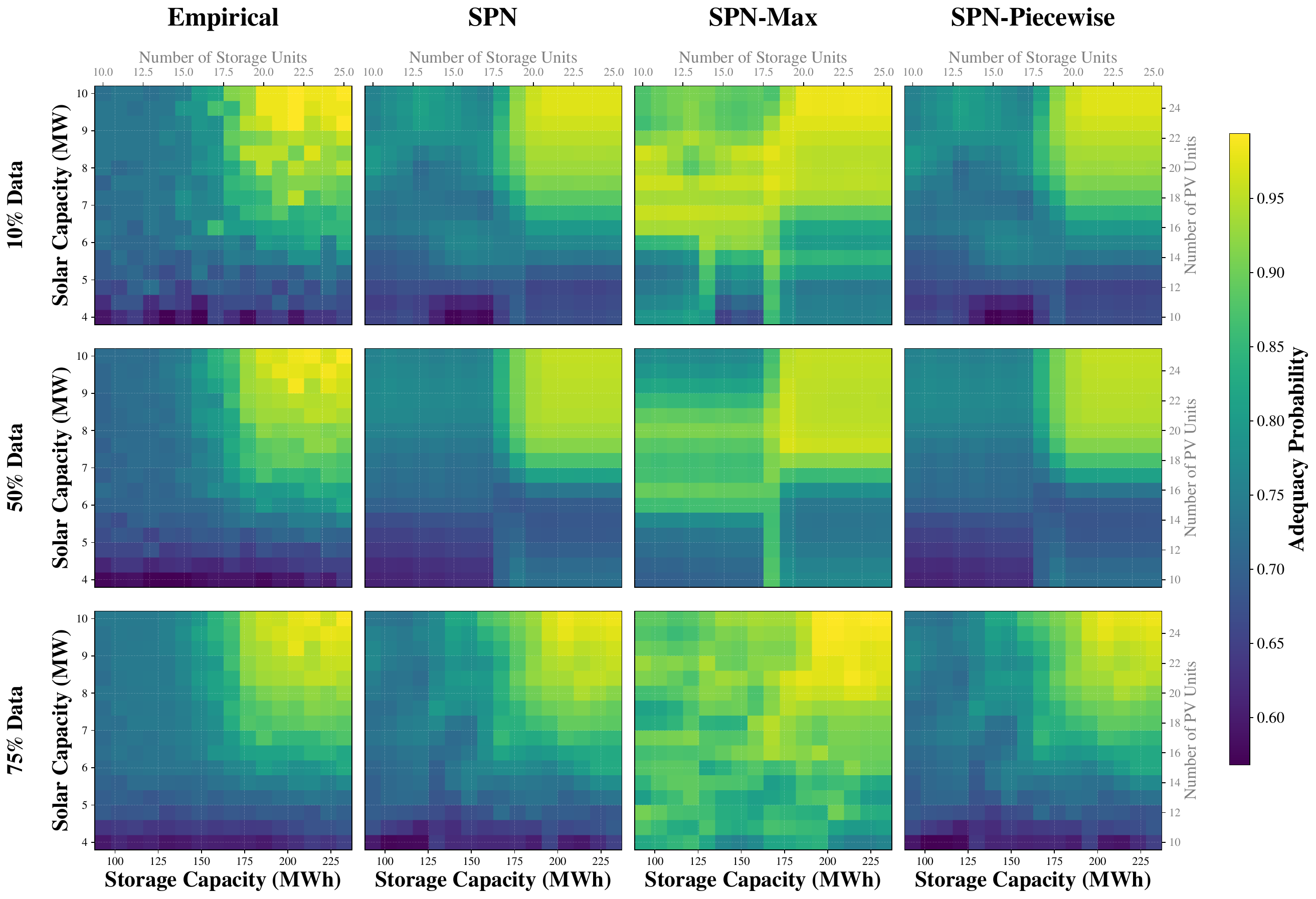}
    \caption{
        \textbf{Comparison of empirical and SPN--based adequacy probability estimates (SPN, SPN--Max, and SPN--Piecewise).} The Empirical column shows the reference probability estimated from the largest available sample set. The SPN column presents the learned approximation. The SPN--Max column illustrates the first proposed approximation, while the SPN--Piecewise column shows the improved approximation. The SPN--Piecewise method most accurately reproduces the empirical landscape. Although SPN--Max captures the general structure, its error increases with SPN size, whereas SPN--Piecewise mitigates this effect while remaining MILP--compatible.}
    \label{fig:spn_heatmaps_subsample}
\end{figure*}
In \autoref{fig:spn_heatmaps_subsample}, horizontal comparisons assess whether SPN--based methods reproduce the empirical probability landscape under a fixed data regime. Vertical comparisons evaluate how probability estimates evolve and converge as the training dataset grows. Heatmaps within each row should display consistent patterns, with high--probability regions appearing in the same locations across methods, indicating that the SPN correctly identifies configurations associated with high adequacy probability.

The results show that SPN increasingly aligns with the empirical reference as more data become available, reflecting improved density estimation. Comparing SPN (full expectation over paths) with SPN--Max (maximum probability path) provides insight into the network structure: large discrepancies indicate that probability mass is distributed across multiple paths, whereas near--identical patterns suggest reliance on a dominant path. This explains why SPN--Max introduces growing approximation error as network complexity increases. The SPN--Piecewise formulation substantially mitigates this effect. 

\begin{figure*}[tb]
    \centering
    \includegraphics[width=0.97\textwidth]{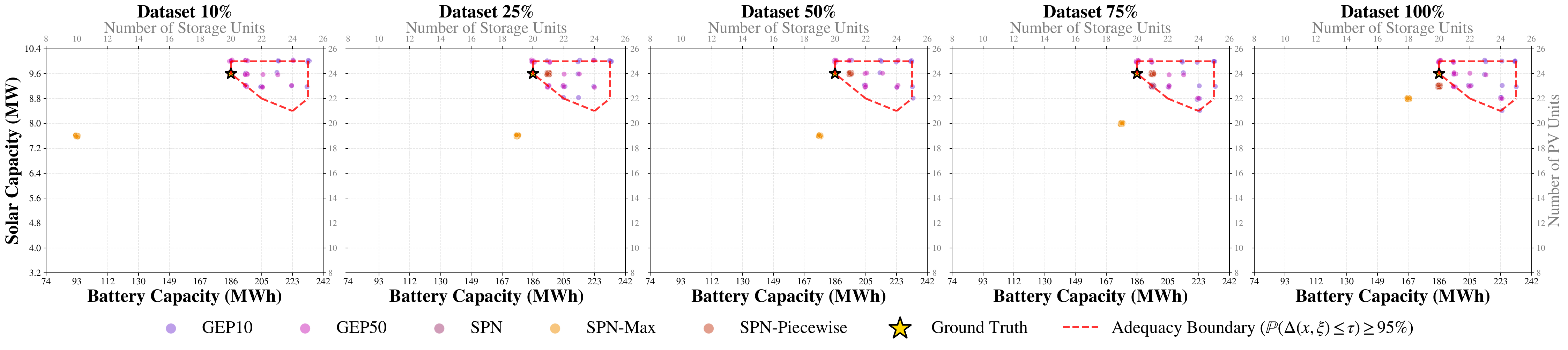}
    \caption{\textbf{Configuration space exploration across data sparsity levels (30 iterations per level).} Point clouds show the configurations selected by each method, with capacity on the primary axes and unit counts on the secondary axes. A gold star denotes the optimal configuration. The red dashed boundary indicates the chance--constraint boundary where $\mathbb{P}(\Delta(x,\xi) \leq \tau) = 1-\varepsilon$ (with $\varepsilon=5\%$); configurations beyond this boundary violate the reliability requirement. Some methods appear safe according to their internal estimates, but select configurations beyond the true safety boundary, revealing calibration overconfidence. A small jitter is added for visualization; solutions cluster around the same configurations.}
    \label{fig:configuration_space_across_percentages}
\end{figure*}

\subsection{Reliability of SPN--Based Investment Planning Under Data Availability Constraints}

This experiment evaluates how the reliability of SPN–based investment planning varies with the availability of training data, building on the analysis in \autoref{sec:spn_fitting}. The objective is to assess whether the learned probabilistic model accurately captures operational feasibility boundaries while keeping the adequacy probability above the threshold, i.e, $\mathbb{P}(\Delta(x,\xi) \leq \tau) \geq 1-\varepsilon$ (with $\varepsilon=5\%$).

We consider five data availability regimes corresponding to $10\%$, $25\%$, $50\%$, $75\%$, and $100\%$ of the full dataset (45,000 realizations). For each regime, 30 independent trials are performed to ensure statistical robustness. Each trial consists of: (i) randomly sampling the dataset, (ii) training an SPN using hyperparameters selected via Bayesian optimization, and (iii) solving the investment problem defined in \eqref{eq:practical_objective}. This setup reflects realistic planning conditions in which only partial operational or synthetic data are available. The full dataset serves as a reference benchmark, providing the most accurate estimate of the adequacy probability. We compare the SPN--based methods against two scenario--based approaches:

\subsubsection{GEP10} Following \eqref{eq:scenario_expansion_og}, an empirical uncertainty set $\hat\Xi$ is constructed with $K = 10$ representative scenarios generated by the simulator and solve:
\begin{align}
    \min_{x}\quad 
    & 
    C^{\mathrm{inv}}(x)\\ \notag
    \text{s.t.}\quad 
    & G(x,\xi) \geq 0,
    \quad
    \quad \forall \xi \in \hat\Xi. \notag
\end{align}

This formulation tends to underestimate risk because of its limited stochastic coverage.

\subsubsection{GEP50} Similar to GEP10, but uses $K = 50$ scenarios. The larger scenario set improves uncertainty representation, generally yielding safer but potentially more expensive solutions.

Each subplot in \autoref{fig:configuration_space_across_percentages} corresponds to a different dataset fraction. The horizontal axis represents the battery capacity (MWh), while the vertical axis shows solar generation capacity (MW). The red dashed contour indicates the adequacy boundary separating feasible from infeasible configurations. The star marks the optimal configuration, obtained from the best possible estimation with the maximum number of simulated samples. This experiment illustrates the progressive improvement of the SPN reliability as the dataset size increases. With limited samples, the SPN captures only coarse structural patterns, which may suffice to identify a feasible region but can lead to imprecise probability estimates. As more data become available, the model refines; however, intermediate data regimes (e.g., $50\%$) may temporarily increase estimation error before full convergence.

The SPN--based approximations depend directly on the quality of the learned density; any modeling error propagates into the MILP formulation. Interestingly, our results suggest that the approximation introduced in the sum--node can induce a conservative bias, effectively acting as a regularizer. In our experiments, this conservative tendency helped maintain decisions within the admissible adequacy threshold across varying data availability levels, which proved essential for reliability preservation.

Comparing the SPN--based approaches with GEP10 and GEP50 reveals the advantages of probabilistic modeling. In our experimental setup, traditional GEP formulations depend critically on the quality and representativeness of the selected scenarios, which are sampled from the available data at each iteration. With a small scenario set (GEP10), stochastic coverage is limited, resulting in high variability across iterations and no reliable guarantee of satisfying the adequacy constraint as shown in \autoref{fig:configuration_space_across_percentages}. Increasing the scenario count in GEP50 improves consistency but still provides no formal assurance that the adequacy threshold is met. In both cases, performance is fundamentally constrained by the representativeness of the sampled scenarios. One can deduce the cost aspects of the decisions, as increased capacity in any axis represents a higher investment. GEP10 exhibits high variability in investment cost, reflecting its sensitivity to scenario selection and the small number of considered realizations ($K$). GEP50 behaves as expected; solutions are generally more expensive than the ground--truth optimum but display lower cost variance. However, the average solution lies close to the adequacy boundary, meaning that modest estimation or simulation errors can push it into the unsafe region. 

In contrast, SPN--based methods tend to produce more conservative decisions, partly due to model--fitting uncertainty and approximation error introduced by the MILP encoding. Importantly, they demonstrate greater consistency across various data availability levels and random samples. This is a clear advantage, since the operation cost $C^{\mathrm{op}}(x,\xi)$ in \eqref{eq:scenario_expansion_cc} is negatively affected by shortfall events. Thus, we argue that TPM--based approaches could provide a more reliable and potentially more cost--effective long--term alternative to traditional scenario--based expansion methods.

\subsection{Training Details}

We trained our SPNs using the \texttt{SPFlow} library \cite{molinaSPFlowEasyExtensible2019}, which provides automatic structure learning and exact inference for tractable probabilistic models. The structure learning relies on hierarchical clustering and recursive variable partitioning to identify conditional independencies in the data. We tuned hyperparameters using \texttt{WandB} \cite{biewaldExperimentTrackingWeights2020} via Bayesian search with 500 trials per dataset, selecting models based on validation log--likelihood. We standardized and discretized features per leaf node type. Individual runs took seconds to minutes, depending on dataset size and model complexity. We used early stopping to mitigate overfitting.
\section{Limitations}

While the proposed framework shows strong potential, several limitations warrant highlighting to facilitate further research. First, the MILP encoding of SPN inference relies on relaxations/linearizations that may introduce approximation error in probability evaluation, which can affect calibration and feasibility near the chance--constraint boundary. Future work will investigate tighter formulations, alternative probabilistic circuits, or hybrid approaches to reduce this error. Second, the quality of probabilistic inference depends on the representativeness of the training data. Real--world datasets may contain bias or insufficient coverage of rare events, thereby affecting the reliability of estimates. Incorporating active learning or uncertainty--aware sampling could improve robustness. Moreover, reliability estimates are exact only \emph{given the quality of the learned model} and do not inherently provide worst--case distributional guarantees. Combining SPN--based modeling with an ambiguity--set outer layer is a promising approach for recovering DRO--style worst--case guarantees. Finally, although the TPM formulation scales more favorably than scenario--based methods, embedding large SPNs within MILP solvers may become computationally demanding as model complexity grows.


\section{Conclusion}

This work presents a probabilistic framework for power system investment planning under uncertainty by embedding Sum–Product Networks (SPNs) within a mixed--integer optimization pipeline. Building on Tractable Probabilistic Models (TPMs), our approach enforces adequacy through chance constraints evaluated via tractable probabilistic queries, avoiding scenario enumeration at solve time. Experiments on a representative generation expansion planning (GEP) case study demonstrated that SPN--based inference can recover meaningful feasibility landscapes across multiple data--availability regimes. The SPN formulation enables reliability--cost trade--offs that are competitive with standard scenario--based baselines. Overall, these findings indicate that TPMs offer a promising path toward scalable, interpretable, data--driven, and reliability--aware expansion planning that integrates uncertainty and optimization into long--term, low--carbon power system planning.


%


\bibliographystyle{IEEEtran}
\bibliography{references}
\vspace{-10pt}

\begin{IEEEbiography}[{\includegraphics[width=1in,height=1.25in,clip,keepaspectratio]{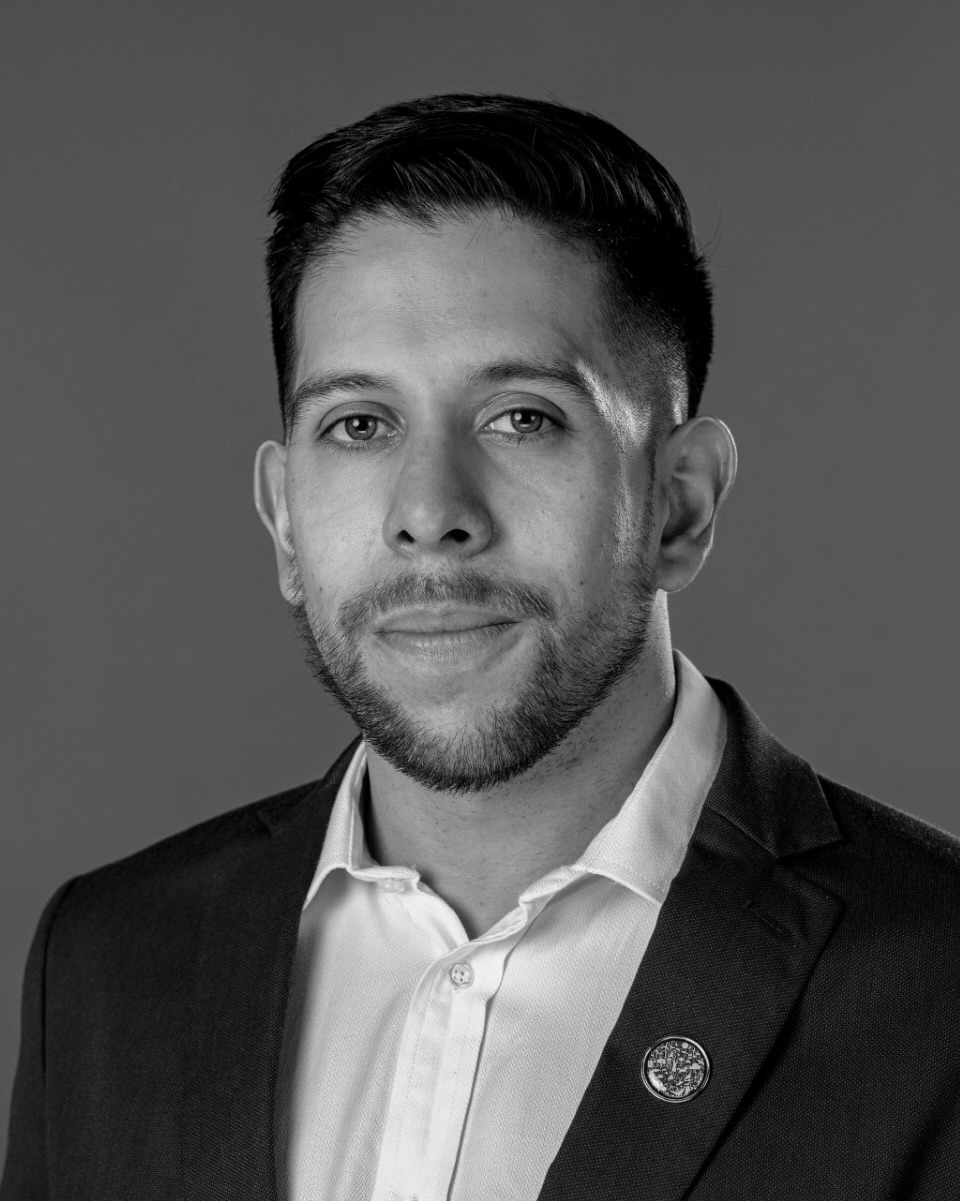}}]{Nicolas Cuadrado}
is a Ph.D. candidate in Machine Learning at the Mohamed bin Zayed University of Artificial Intelligence (MBZUAI), Abu Dhabi, United Arab Emirates. He received the B.Sc. degree in Electronics Engineering from Universidad Nacional de Colombia, Colombia, and the M.Sc. degree in Machine Learning from MBZUAI. His research interests include efficient machine learning, model pruning and compression, reinforcement learning, and AI for sustainable energy systems. He won the GovTech Prize at the World Government Summit in 2018 for the project ``Cycle.'' He has served as a Teaching Assistant at MBZUAI and as a reviewer for ICLR, NeurIPS, AISTATS, and ICML.
\end{IEEEbiography}

\begin{IEEEbiography}[{\includegraphics[width=1in,height=1.25in,clip,keepaspectratio]{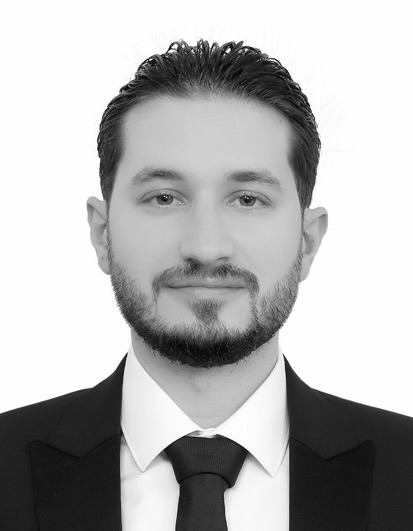}}]{Mohannad Takrouri}
is a Ph.D. candidate in Machine Learning at Mohamed bin Zayed University of Artificial Intelligence (MBZUAI), Abu Dhabi, UAE. He received the M.Sc. degree in Mechatronics Engineering from the American University of Sharjah, UAE, in 2015, and the B.Sc. degree in Mechatronics Engineering from Palestine Polytechnic University, Hebron, Palestine, in 2013. His research interests span intelligent energy systems, smart grid optimization, and reinforcement learning–based control for demand--side management. His work in mechatronics systems and control includes nonlinear friction identification and compensation in voice--coil servomotors, adaptive and resonant control techniques, and real--time implementation using dSPACE platforms. He has also contributed to the modeling and monitoring of large--scale solar photovoltaic systems and to the integration of energy storage.
\end{IEEEbiography}

\begin{IEEEbiography}[{\includegraphics[width=1in,height=1.25in,clip,keepaspectratio]{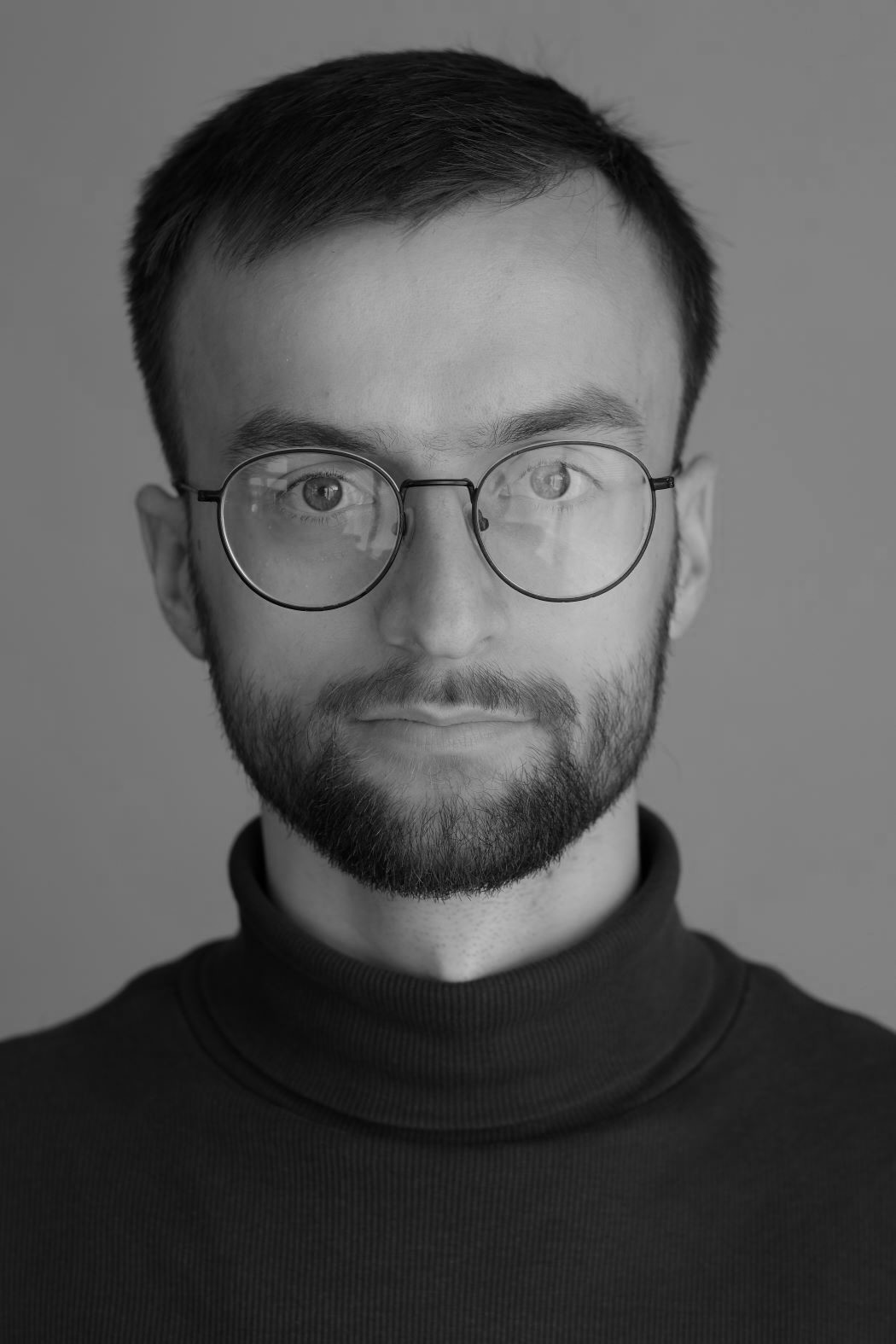}}]{Jiří Němeček}
is a Ph.D. student at the Artificial Intelligence Center, Faculty of Electrical Engineering, Czech Technical University in Prague, Czech Republic. His research focuses on explainability, fairness, and trustworthy artificial intelligence, with work on optimization--based methods for bias detection and counterfactual explanations. He received the M.Sc. degree in Artificial Intelligence from CTU in Prague, graduating with honors, where his thesis explored the application of mixed--integer programming to machine learning for explainable models. He also holds a B.Sc. degree in Knowledge Engineering from CTU in Prague. His publications include contributions to ICLR and KDD on explainable and fair AI methods.
\end{IEEEbiography}

\begin{IEEEbiography}[{\includegraphics[width=1in,height=1.25in,clip,keepaspectratio]{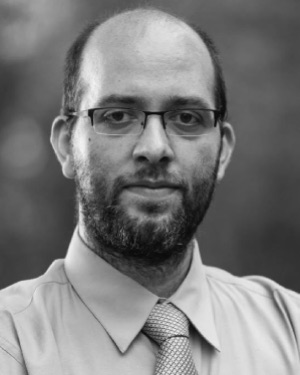}}]{Martin Takáč}
received the B.S. and M.S. degrees in mathematics from Comenius University, Slovakia, in 2008 and 2010, respectively, and the Ph.D. degree in mathematics from the University of Edinburgh, U.K., in 2014. He is currently an Associate Professor at the Mohamed bin Zayed University of Artificial Intelligence (MBZUAI) in the United Arab Emirates. Before joining MBZUAI, he was an Associate Professor in the Department of Industrial and Systems Engineering at Lehigh University, where he has been employed since 2014. His current research interests include designing and analyzing algorithms for machine learning, AI for science, understanding protein--DNA interactions, and using ML for energy. He received funding from various U.S. National Science Foundation programs (including through a TRIPODS Institute grant awarded to him and his collaborators at Lehigh, Northwestern, and Boston University) and was recently awarded a grant with the Weizmann Institute of Science. He served as an Associate Editor for Mathematical Programming Computation, Journal of Optimization Theory and Applications, and Optimization Methods and Software, and an Area Chair for ICLR and AISTATS. He also serves as an Area Chair for ICML and NeurIPS.
\end{IEEEbiography}

\begin{IEEEbiography}[{\includegraphics[width=1in,height=1.25in,clip,keepaspectratio]{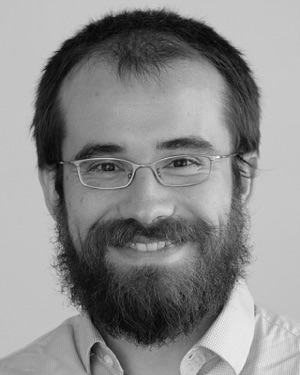}}]{Jakub Marecek}
received the first two degrees in computer science from Masaryk University, Brno, Czech Republic, in 2006 and 2009, respectively, and the Ph.D. degree in computer science from the University of Nottingham, Nottingham, U.K., in 2012, with a focus on mathematical optimization. He was also early employee with two start--ups; worked in R\&D of ARM Ltd., at the University of Edinburgh, Edinburgh, U.K.; at the University of Toronto, Toronto, ON, Canada; at IBM Research – Ireland; and at the  University of California, Los Angeles, Los Angeles, CA, USA. He has been a tenured faculty member at the Czech Technical University in Prague, Czech Republic, since 2020. His research interests include the design and analysis of algorithms for optimisation and control problems across a range of application domains.
He has received awards for his work for transmission system operators while at IBM Reserach, for his work in energy trading at CEZ a.s., as well as for his work in Intelligent Transportation Systems.   
He serves as an associate editor at the IEEE Open Journal of Intelligent Transportation Systems
and as an
Area Chair for ICLR.
\end{IEEEbiography}

\vfill

\end{document}